\documentclass[9pt,twocolumn,twoside]{pnas-new}

\articletype{CLASSIFICATION}

\templatetype{pnasresearcharticle} 

\begin{document}

\title{Principles of frugal inference and control}

\fancyhead[LO]{}
\fancyhead[RO]{}
\fancyhead[LE]{}
\fancyhead[RE]{}

\author[a]{Itzel Olivos-Castillo}
\author[b,c]{Paul Schrater}
\author[a,d,e,f,g,h,i,*]{Xaq Pitkow}


\affil[a]{Department of Computer Science, Rice University}
\affil[b]{Department of Computer Science, University of Minnesota}
\affil[c]{Department of Psychology, University of Minnesota}
\affil[d]{Neuroscience Institute, Carnegie Mellon University}
\affil[e]{Department of Machine Learning, Carnegie Mellon University}
\affil[f]{Department of Neuroscience, Baylor College of Medicine}
\affil[g]{Department of Electrical and Computer Engineering, Rice University}
\affil[h]{Center for Neuroscience and Artificial Intelligence, Baylor College of Medicine}
\affil[i]{NSF AI Institute for Artificial and Natural Intelligence}

\leadauthor{Pitkow}

\significancestatement{For biological and artificial systems, inferring hidden variables from noisy observations consumes substantial computational resources. Yet most theories of control under uncertainty implicitly assume that inference is free. Here, we treat information gained through inference as a resource that must be optimized alongside utility. When information is costly, optimal behavior fundamentally changes: agents deliberately leave some uncertainty unresolved and compensate through physical effort. In this `frugal' regime, optimal solutions are no longer unique but form a structured family of equally effective strategies that differ in how internal computations and external movement are coordinated. This research identifies general principles of computational efficiency for inference and control, offering insight into how brains and machines could achieve reliable behavior under tight computational constraints.}

\authorcontributions{All authors jointly conceived the study and mathematical framework, and contributed analyses. IOC implemented the simulations, performed most mathematical analyses, and wrote the first draft of the manuscript. IOC and XP edited the manuscript. XP and IOC acquired funding. XP and PS supervised the project.}

\correspondingauthor{\textsuperscript{*}To whom correspondence should be addressed. E-mail: xaq@cmu.edu}

\keywords{Control $|$ resource efficiency $|$ uncertainty $|$ ...}

\begin{abstract}
A central challenge for intelligent agents in an uncertain world is striking the right balance between utility maximization and resource use—not only for external movement but also for internal computation. Existing theories of control under uncertainty typically treat inference as cost-free, despite the substantial computational and energetic burden it imposes in both artificial and biological systems. To remedy this problem, we introduce a novel variant of the POMDP framework in which the information acquired through inference is treated as a resource that must be optimized alongside utility. Solving a local linear–Gaussian approximation of the resulting problem reveals three general principles of resource-efficient control.  First, when information is costly, inference shifts from a Bayes-optimal (lossless) compression of the past to a lossy regime that strategically leaves some uncertainty unresolved to optimize resource use. Second, relaxing exact Bayesian inference creates a manifold of equivalent solutions, reflecting multiple ways to combine imperfect inference with compensatory control. This flexibility can be used to meet additional objectives or constraints without sacrificing performance on the original task. Third, beyond goal attainment, control can be leveraged to counteract estimation errors and steer the system into regimes where representation costs are lower. We empirically demonstrate that these principles generalize beyond the local linear–Gaussian approximation, enabling the solution of nonlinear control problems such as pole balancing and drone stabilization. Together, these results establish a framework for rational computation that extends existing approaches to information-constrained decision-making and offers normative insight into how brains and machines can achieve effective behavior under tight computational constraints.
\end{abstract}

\dates{This manuscript was compiled on \today}

\maketitle
\thispagestyle{firststyle}
\ifthenelse{\boolean{shortarticle}}{\ifthenelse{\boolean{singlecolumn}}{\abscontentformatted}{\abscontent}}{}

\Firstpage

Intelligent behavior requires acting under uncertainty. Since task-relevant variables in natural environments are often hidden and constantly changing, intelligent agents—biological and artificial—must infer unobserved aspects of the world from noisy sensory data to guide their actions. A powerful framework for studying this problem is the partially observable Markov decision process (POMDP), which formalizes how inferences over hidden states support long-term decision making. Optimal behavior in a POMDP is achieved by selecting actions that maximize expected utility with respect to a Bayes-optimal posterior probability: a `belief state' that summarizes the relevant history of past observations and actions \cite{kurniawati2022partially}. Building and maintaining exact beliefs is computationally demanding, especially in continuous, high-dimensional settings. Sampling-based methods \cite{ross2011bayesian, lim2023optimality} and variational techniques \cite{watter2015embed, ha2018world, igl2018deep, hafner2019learning, hafner2025mastering} are common approaches to belief updating in complex sequential decision making problems; however, their computational demands scale steeply with state dimension, model complexity, and desired accuracy, often requiring large particle sets, expressive variational families, or extensive optimization. This computational burden often exceeds the capacity of real-world agents. For example, planetary rovers operate in environments that expose electronics to extreme conditions, which requires radiation-hardened processors whose robustness comes at the  \Parasplit expense of processing speed \cite{nasa_perseverance_computer}. Implantable brain–computer interfaces must adhere to severe limits on energy consumption, heat dissipation, and inference latency to ensure safety, stability, and real-time responsiveness \cite{karageorgos2020hardware}. Similarly, mass-market robots—such as autonomous vacuums, pool cleaners, and lawnmowers—are constrained to modest onboard computation, as incorporating specialized parallel computing hardware would undermine commercial viability \cite{vaussard2014lessons}. Despite this reality, most theoretical treatments of sequential decision making under uncertainty, whether framed as planning or feedback control, treat inference as a fixed subroutine, assuming that agents can always perform Bayes-optimal (or near optimal) belief updating whenever uncertainty is present. This assumption obscures a fundamental question: When resources are limited, how should an agent decide how much uncertainty to resolve in order to balance utility gains against computation?

The resource-utility trade-off was first recognized in Simon’s work on bounded rationality \cite{simon1955behavioral} and later formalized in bounded optimality and meta-reasoning frameworks \cite{russell1994provably, russell1991principles, cox2005metacognition, horvitz2013reasoning}. Building on this foundation, information-theoretic approaches \cite{tishby2010information, rubin2012trading, ortega2013thermodynamics, ortega2015information} have clarified how sequential decision-making can be optimized under resource constraints in fully observable settings, in ways that are independent of hardware specifics and implementation details. These frameworks have led to scalable algorithms \cite{tsiotras2021bounded} and extensions that account for model uncertainty \cite{grau2016planning}. Furthermore, empirical studies have shown that information-theoretic constraints can account for the emergence of heuristics in human decision making \cite{binz2022heuristics}, explain how people balance reward maximization with cognitive effort \cite{lancia2023humans}, and characterize the pervasive coexistence of habitual and controlled responses across many tasks \cite{moskovitz2022unified}. However, extending these ideas to partially observed environments remains an open challenge. Most studies focus on explaining cognitive-level behavior without pursuing the computational implementations \cite{kool2018planning, ho2022people, ongchoco2024people}. Other works showcase practical implementations in complex tasks but do not seek to derive theoretical insights \cite{pedram2021rationally, mazzaglia2021contrastive, pacelli2022robust}. The few works that both aim to derive principles and demonstrate applicability are restricted to single-step decision-making problems \cite{howard2007information, gershman2010neural, genewein2015bounded, schmid2024bounded} or settings that solely address perception or communication constraints \cite{sims2003implications, tatikonda2004control, susemihl2014optimal}. To date, understanding how the computational burden of inference reshapes optimal behavior in complex, partially observed control tasks remains an open problem, explored primarily at a conceptual level \cite{horvitz2013reasoning, gershman2015computational, grujic2022rational}.

To address this gap, we introduce a novel variant of the POMDP framework in which inference is an explicit object of optimization. This adds a new dimension to the classic trade-off between goal achievement and physical effort: the internal cost of resolving state uncertainty. To reveal the structure of this new computationally constrained regime, we solve a local linear–Gaussian approximation of the general problem and thoroughly characterize the solutions. These take the form of frugal strategies that jointly specify how previous evidence should be integrated and transformed into actions, optimally balancing task performance, control effort, and the information acquired through inference. Our study reveals three principles of computational efficiency. First, when information is costly, inference undergoes a phase transition from Bayes-optimal inference, which provides a lossless compression of the past, to a lossy regime in which agents strategically leave some epistemic uncertainty unresolved to optimize resource use. Second, beyond this transition, frugal solutions are no longer unique. Instead, agents can combine imperfect inference with compensatory control in multiple ways, forming a structured family of equally effective strategies. This multiplicity provides flexibility, allowing agents to accommodate new objectives or constraints without sacrificing performance on the original task. Third, beyond goal attainment, control can be leveraged to counteract estimation errors and simplify inference, the latter by steering the system toward states that are less variable and thus require less uncertainty to be resolved. We validate the extension of these principles to nonlinear control problems, including pole balancing and drone stabilization. By treating inference as a regulated process rather than a fixed subroutine, this work extends existing frameworks for information-constrained decision-making in two key ways: it generalizes these frameworks from fully observed to partially observed settings and it reconceptualizes control not only as a means of achieving external goals but also as a mechanism for decreasing representation cost. More broadly, our results establish a foundation for a new type of rational computation that both brains and machines could use to achieve effective behavior under tight computational constraints.

\section*{Results}

\subsection*{Control when information is costly}

\begin{figure*}[t!]
\centering
\includegraphics[scale=0.25]{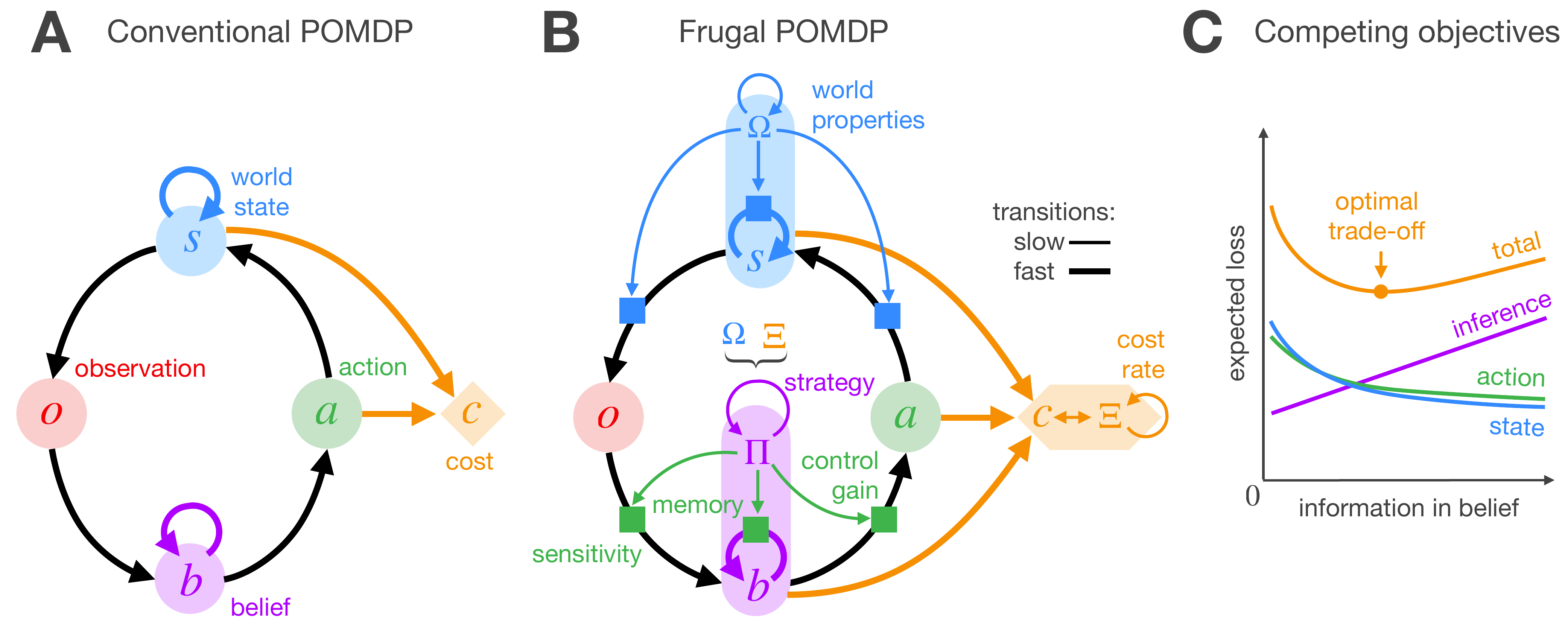}
\caption{Structure of computationally constrained control. \textbf{A)} Conventional POMDP. The agent interacts with a hidden world state over time, receiving noisy observations, taking actions that change the state, and incurring costs based on the action taken and the resulting next state. Minimizing cumulative costs requires managing state uncertainty. To address this, the agent builds and updates a belief over the hidden state that aims to fully summarize previous evidence (the history of past observations and actions). \textbf{B)} Frugal POMDP. The agent pays for the information that beliefs encode about hidden world states. To balance this internal cost against state and action costs, the agent computes a strategy $\Pi$ that dictates how to integrate new evidence and how to transform the resulting beliefs into actions. To compute this strategy, the agent considers the properties of the world, $\Omega$, and the penalty parameters of the cost functions, $\Xi$, factors that we assume are fully observable and change slowly. \textbf{C)} Optimal trade-off. State and action costs decrease as the belief encodes more information about the hidden state. However, when the cost of storing information is considered, the agent can improve overall performance by tolerating more state and action costs if doing so saves enough bits in the inference.}
\label{fig:F1}
\end{figure*}

In a conventional POMDP (Figure \ref{fig:F1}A), the set of task-relevant variables (world state $s_t$) is hidden. To mitigate this uncertainty, the agent uses probabilistic inference to build and update a belief $b_t$ that summarizes the history of previous observations $o_{\leq t}$ and actions $a_{<t}$. The more information the belief encodes about the hidden state, the more purposeful and effective the agent's actions can be. However, every bit of information gained through inference comes at the cost of time, computation, memory, and energy. To investigate how accounting for this burden reshapes rational behavior, we develop a frugal variant of the POMDP framework (Figure \ref{fig:F1}B) in which belief updating is treated as a regulated process rather than a fixed subroutine. Planning in this POMDP therefore entails jointly optimizing inference and control (action selection) to minimize the expected cost of deviating from target states, $C_s(s_t)$, the cost of acting, $C_a(a_t)$, and the cost of reducing uncertainty via inference, $C_b(s_t,  b_t)$:
\begin{equation}
\mathbb{E}\left[\sum_t C_s(s_t)  + C_a(a_t)  + C_b(s_t,  b_t)\right]
\label{eq:loss_fun}
\end{equation}
To balance these competing objectives (Figure \ref{fig:F1}C), we assume that the agent has exact knowledge of the cost functions, $\Xi=\{C_s(\cdot), C_a(\cdot), C_b(\cdot)\}$, and the world properties, $\Omega$, defining state transitions and observation generation. We further assume that $\Xi$ and $\Omega$ change slowly over time, allowing the agent to gradually adapt its parameters for extended periods of stable control at equilibrium. During the adaptation phase, the agent computes a frugal strategy $\Pi=\{\pi^b, \pi^a\}$ that jointly specifies how past evidence is integrated into beliefs—via an inference process parameterized by $\pi^b$ (internal, computational actions)—and how those beliefs are mapped into external actions through a control policy $\pi^a$. Following adaptation, the agent enters a long period of stable control, interacting with the world using the beliefs and actions derived from the strategy $\Pi$, that is, $ b_t = f\left(o_{\leq t}, a_{<t}; \pi^{b}\right)$ and $a_t = g\left(b_{t}; \pi^{a}\right)$. 

\subsection*{An interpretable approximation}

In general POMDPs, the parameters defining the strategy that optimally balances state, action, and inference costs can be complex, such as the weights of a recurrent neural network. However, when the dynamics are locally approximated by a linear-Gaussian model and the reward function by a quadratic form, the problem becomes tractable and the solution interpretable. Under this approximation, the hidden state evolves according to stochastic linear dynamics, $s_t = D s_{t-1} + E a_{t-1} + w_{t-1}$, and observations are linear, noisy versions of the hidden state $o_t = s_t + v_t$. Here, the dynamics matrix $D $ captures how unstable the state is, the input gain $E $ characterizes actuator responsiveness, $a_{t-1} $ is the action taken by the agent, $w_{t-1} $ is additive white Gaussian process noise with isotropic covariance $Q $, and $v_t $ is additive white Gaussian observation noise with isotropic covariance $R $. The cost function takes the quadratic form $\mathbb{E}\left[\sum_t s_t^\top C_s s_t  + a_t^\top C_a a_t  + C_b \ \mathcal{I}\left(s_t; b_t\right)\right]$, with the mutual information between states and beliefs, $\mathcal{I}(\cdot)$, capturing the cost of reducing uncertainty via inference. 

Without our added internal inference cost, $C_b \ \mathcal{I}\left(s_t; b_t\right)$, the linear–quadratic–Gaussian (LQG) approximation allows for an analytical solution with theoretical guaranties—the LQG controller—which combines Bayes-optimal inference via a Kalman filter with optimal control via a linear quadratic regulator \cite{bertsekas2012dynamic}. Due to these advantages, the LQG approximation is widely used to model local dynamics in complex, nonlinear, continuous-state control problems, enabling faster learning \cite{levine2014learning}, facilitating adaptive control on hardware with limited onboard computation \cite{gao2025enabling}, and, more broadly, serving as a testbed for uncovering fundamental principles in control theory \cite{tang2023analysis}, reinforcement learning \cite{hu2023toward}, and neuroscience \cite{todorov2005stochastic, susemihl2014optimal, boominathan2022phase}. 

We, too, capitalize on the analytical tractability of the LQG approximation, but our solution does not rely on the LQG controller. In Methods, we detail the technical considerations that arise when the inference process is subject to optimization, and provide a comprehensive description of our approach for addressing them. Briefly, we solve the planning problem numerically at equilibrium, where the expected total cost is entirely defined by a steady-state covariance matrix $\Sigma$ that captures the dependencies among states, observations, beliefs, and actions. The result is an interpretable strategy that specifies how much of the past should be remembered and how beliefs should be translated into actions, achieving an optimal trade-off among state, action, and inference costs.

\subsection*{Principles of resource-efficient control} 

\subsubsection*{Spend when it counts}

\begin{figure*}[t!]
\centering
\includegraphics[scale=0.25]{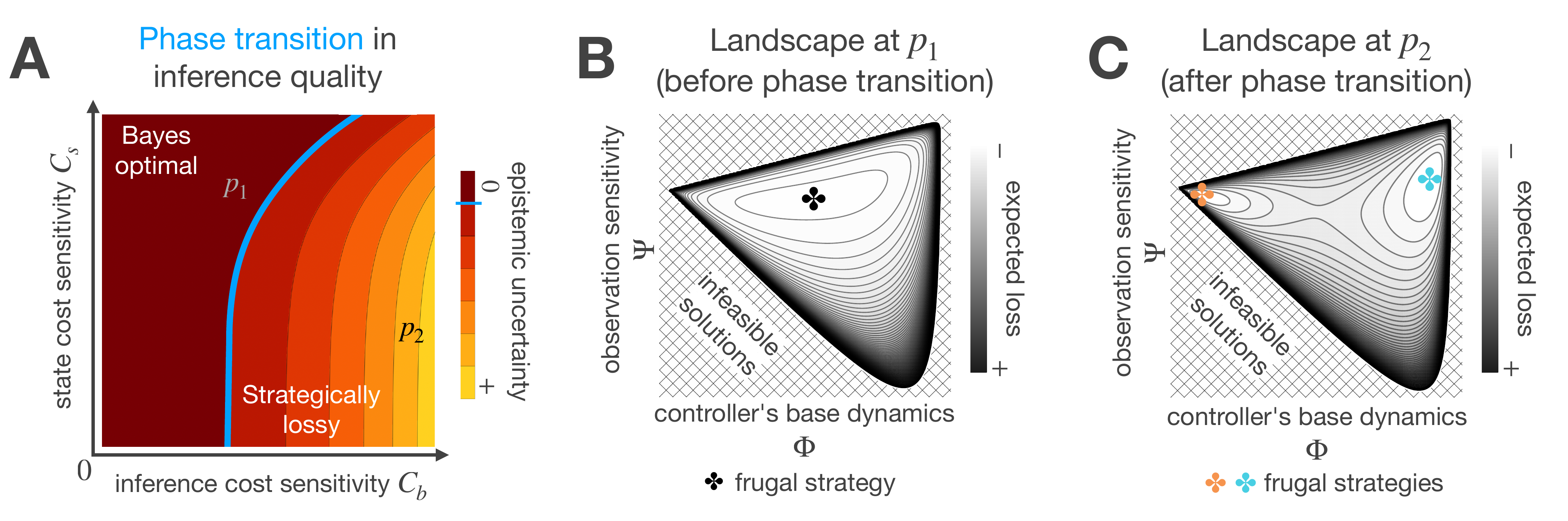}
\caption{Parameter space for frugal inference. \textbf{A)} Phase transition in the optimal inference strategy. The penalties $C_s$ and $C_b$, which determine the relative importance of minimizing state deviations and reducing information use, set a threshold (blue line) beyond which the benefits of Bayes-optimal (lossless) inference saturate. Markers $p_1$ and $p_2$ indicate parameters at which the optimization landscapes of Plots B and C are defined. \textbf{B)} Optimization landscape before the phase transition. The optimization landscape of the planning problem is convex when the agent relies on lossless inference. \textbf{C)} Optimization landscape after the phase transition. When the agent leaves some epistemic uncertainty unresolved, the optimization landscape has multiple global minima. The multiple solutions achieve statistically equivalent performance but differ in how the agent integrates new evidence, offsets estimation errors, and generalizes to novel settings.}
\label{fig:F2}
\end{figure*}

Unburdened by any computational constraints, the optimal solution to control under uncertainty involves selecting actions based on beliefs derived from a Bayes-optimal inference process that mitigates reducible (epistemic) uncertainty and accurately quantifies irreducible (aleatoric) uncertainty, effectively providing a lossless compression of past observations. However, for agents operating under computational limitations, the demands of lossless inference may exceed processing capacity. In such cases, conserving computational resources—by reducing the information distilled from previous evidence—emerges as an additional objective that competes against optimizing goal achievement (state cost) and motion effort (action cost). The solution to this computationally constrained control problem is a frugal strategy that jointly specifies how past evidence is integrated as a function of task demands, environmental reliability, and resource availability, and how the resulting beliefs are transformed into actions that pursue goal attainment, compensate for estimation errors and/or reduce representational cost. Figure \ref{fig:F2}A illustrates this behavior in a scalar task. The agent invests in lossless inference only when it is affordable—meaning the cost per bit of information gained through inference, $C_b$, is low—or essential, which occurs when the cost per unit of deviation from the target state, $C_s$, is high. Otherwise, the agent relies on a lossy inference approach that leaves some epistemic uncertainty unresolved to optimize resources. This phase transition in inference quality, from lossless to lossy, coincides with a shift in the optimization landscape of the control problem from convex (Figure \ref{fig:F2}B) to having multiple global minima (Figure \ref{fig:F2}C).

\subsubsection*{Adaptability begins when perfection ends}\label{subsubPrinciple2}

\begin{figure*}[t!]
\centering
\includegraphics[scale=0.25]{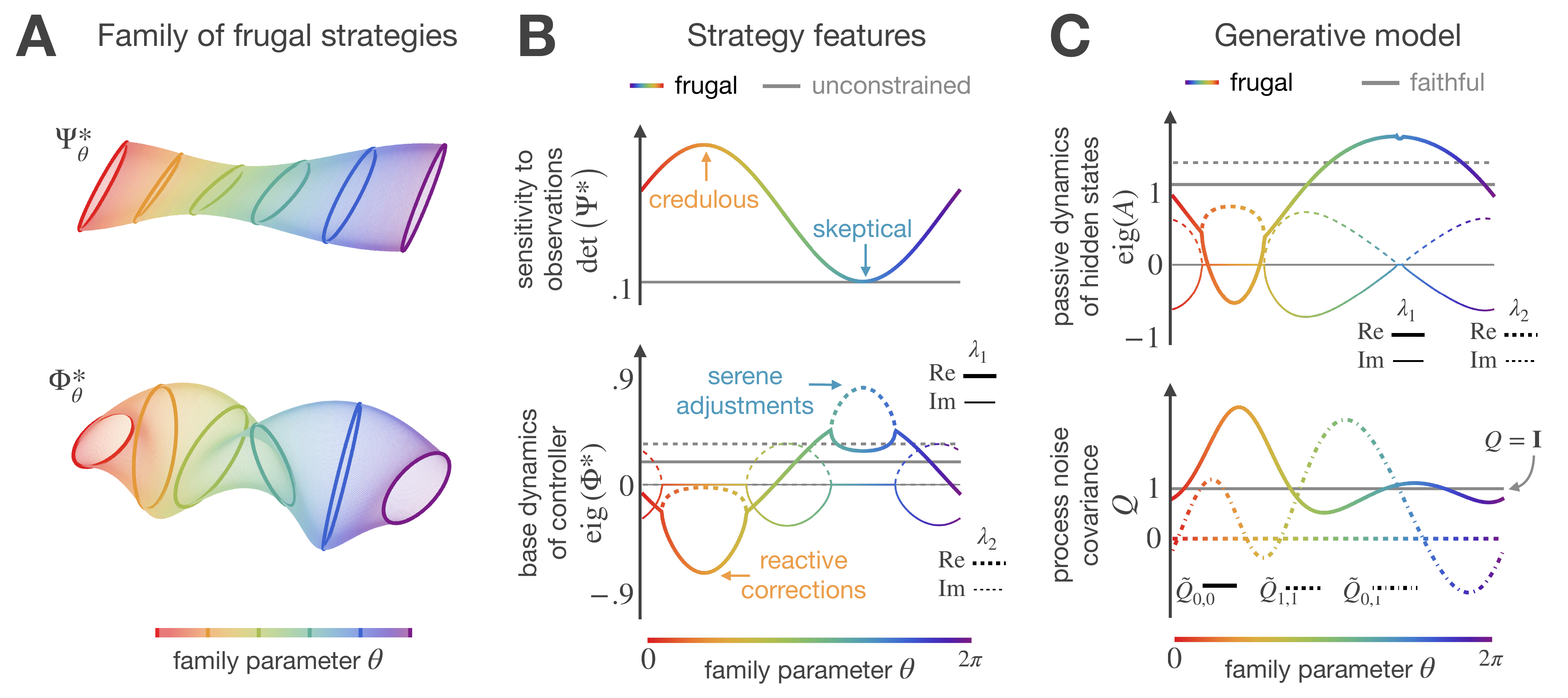}
\caption{Family of frugal strategies. \textbf{A)} Graphical representation of optimized strategies. For $2$-dimensional control tasks, the solutions are described by 2$\times$2 matrices of observation sensitivity $\Psi$ and controller's base dynamics $\Phi$. Here, these matrices are visualized by how they transform a unit circle into an ellipse. Members of the solution family are related by an orthogonal transformation that is fully defined by a free angle $\theta$ (depicted by hue). For each color there is a pair of ellipses for the surfaces of $\Psi$ and $\Phi$, representing a strategic combination of lossy inference and error-aware control. \textbf{B)} Strategy features. The family members differ in how the agent integrates new evidence and offsets estimation errors. For instance, the strategies prioritizing observations over predictions (credulous) require controllers that frequently change the direction of motion (reactive). In contrast, the strategies that prioritize predictions over observations (skeptical) rely on controllers that correct deviations with gradual, smooth movements (serene). The combination of inference and control that solves the control problem when inference is free is shown in gray. \textbf{C)} Generative model. To save bits in the inference, the agent makes deliberately mistaken assumptions about the world. Some strategies model the stochasticity in the transition dynamics as stable oscillations with high process noise (orange-hued members); others explain this randomness as low process noise in a volatile environment (blue-hued members). The ground truth properties are shown in gray.}
\label{fig:F3}
\end{figure*}

When the agent relies on lossy beliefs, the solution to the computationally constrained control problem becomes a family of frugal strategies. Intuitively, this occurs because, while there is only one way to achieve Bayes-optimality, there are many ways to make mistakes that leave some epistemic uncertainty unresolved. Subsection \textit{Recovering the complete solution family} provides the mathematical justification of this intuition. We also demonstrate that the solution family has structure: its members are related by a free orthogonal transformation, which allows the recovery of the entire family from a single solution. This transformation manifests itself as a reflection for scalar tasks, which accounts for the two global minima observed in Figure \ref{fig:F2}C. However, in the multivariate context, the transformation gives rise to countless combinations of perception and mobility. To demonstrate this, Figure \ref{fig:F3}A visualizes the characteristics of a family solving a $2$-dimensional task. 

All family members yield statistically equivalent state, action, and internal inference costs; however, due to differences in their temporal structure, they differ in how the agent integrates new evidence and offsets estimation errors. The controller's input-output form, $a_t = \Phi a_{t-1} + \Psi o_t$, allows us to explore these differences. Here, the controller's base dynamics $\Phi$ defines how the actions evolve without new evidence, while the observation sensitivity $\Psi$ determines which dimensions can be attenuated or magnified without compromising eventual goal attainment. As Figure \ref{fig:F3}B illustrates, the solutions allow multiple distinct combinations of lossy inference and error-aware control. For example, the orange-hued strategies combine ``credulous inference'' with ``reactive control,'' which means they give new observations more credence compared to predictions, so they continuously reverse the direction of motion to reactively correct estimation errors. In contrast, the blue-hued solutions pair ``skeptical inference'' with ``serene control,'' tending to disbelieve observations and instead favor prior predictions, and they rely on a gradual correction of major deviations. The different ways family members integrate new evidence can be understood as lossless inference based on a mistaken world model (Figure \ref{fig:F3}C). For example, the credulous and reactive strategy models the stochasticity in the world as stable oscillations coupled with high process noise. Conversely, the skeptical and serene solution interprets stochasticity as low process noise in a volatile environment.

Frugal inference thus offers a powerful new perspective on decision-making. Relaxing strict task optimality reveals orthogonal transformations that define a free design subspace, which agents can exploit to accommodate additional objectives or constraints without sacrificing baseline performance. In reinforcement learning, this property is particularly advantageous: it can enable rapid adaptation to distributional shifts, model misspecification, and computational limitations without destabilizing control. Adaptability thus emerges not from solving the full optimization problem anew, but from principled navigation within a manifold of equivalent solutions.

\subsubsection*{Thinking less, moving more}

\begin{figure*}[t!]
\centering
\includegraphics[scale=0.25]{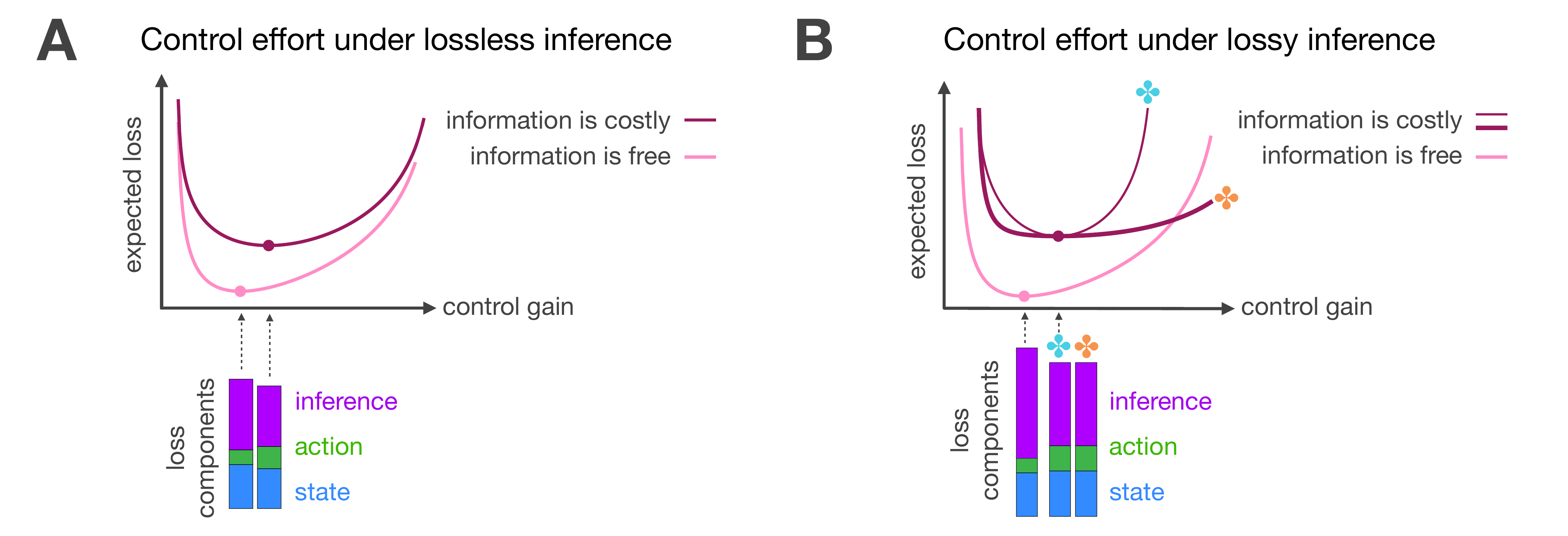}
\caption{Trade-off between control effort and inference cost in a scalar task. The best control gain with costly information is higher than the best control gain when information is free. This additional effort serves two purposes, depending on inference quality, as shown here: \textbf{A)} Additional action cost when using lossless inference. The agent applies a strong control gain to decrease state variance; this approach indirectly lowers inference cost by reducing the variability that previous evidence has to explain. \textbf{B)} Additional action cost when using lossy inference. The agent applies a strong control gain to offset the estimation errors arising from unresolved uncertainty.}
\label{fig:F4}
\end{figure*}

When information is costly ($C_b > 0$), agents optimally trade control effort for reductions in inference cost. This additional effort serves two complementary roles. When lossless beliefs are required to avoid catastrophic errors (Figure \ref{fig:F4}A), the agent applies stronger control to make the problem easier to infer, steering the system toward states that are less variable and thus require less uncertainty to be resolved. Conversely, when lossless compressions of the past are no longer required (Figure \ref{fig:F4}B), increased control compensates for the estimation errors introduced by cheaper, approximate beliefs.

For frugal agents, control serves not only goal attainment, but also error correction and the simplification of inference. This perspective departs fundamentally from existing approaches to information-constrained decision-making. In uncontrollable (yet stable) settings, where inference is the sole mechanism for managing uncertainty, frugal inference closely parallels Efficient Coding \cite{simoncelli2001natural}: representations adapt to the statistical structure of the environment to optimize limited resources, trading representational fidelity against cost. The picture changes qualitatively when control becomes available. First, control objectives induce a task-dependent notion of relevance, as different state dimensions contribute unequally to successful goal attainment; consequently, frugal inference preserves fidelity primarily along task-relevant directions while distorting irrelevant ones. Second, uncertainty can be mitigated not only internally, through improved inference, but also externally, through actions that steer the system toward states that are easier to estimate. By framing control as a mechanism for goal attainment, error correction, and simplifying inference, our framework departs from rational inattention \cite{sims2003implications}, information-constrained bounded rationality \cite{ortega2015information}, and active inference \cite{friston2017active}, where control is typically optimized given resource-constrained beliefs but does not explicitly lessen the intrinsic complexity of uncertainty reduction.

\subsection*{Empirical validation in nonlinear control tasks}
\label{secCases}

We selected two nonlinear tasks to illustrate the generality of the principles of computational efficiency we derived. In both cases, the agents are simplified models of practical machines operating under tight computational budgets. To study these systems, we first derive a linear-quadratic-Gaussian approximation around the target state. Next, we compute the frugal strategies using the approach described in Methods. Finally, we assess performance by executing the strategies in the original nonlinear system.

\subsubsection*{Frugal cart pole}

\begin{figure*}[t!]
\centering
\includegraphics[scale=0.25]{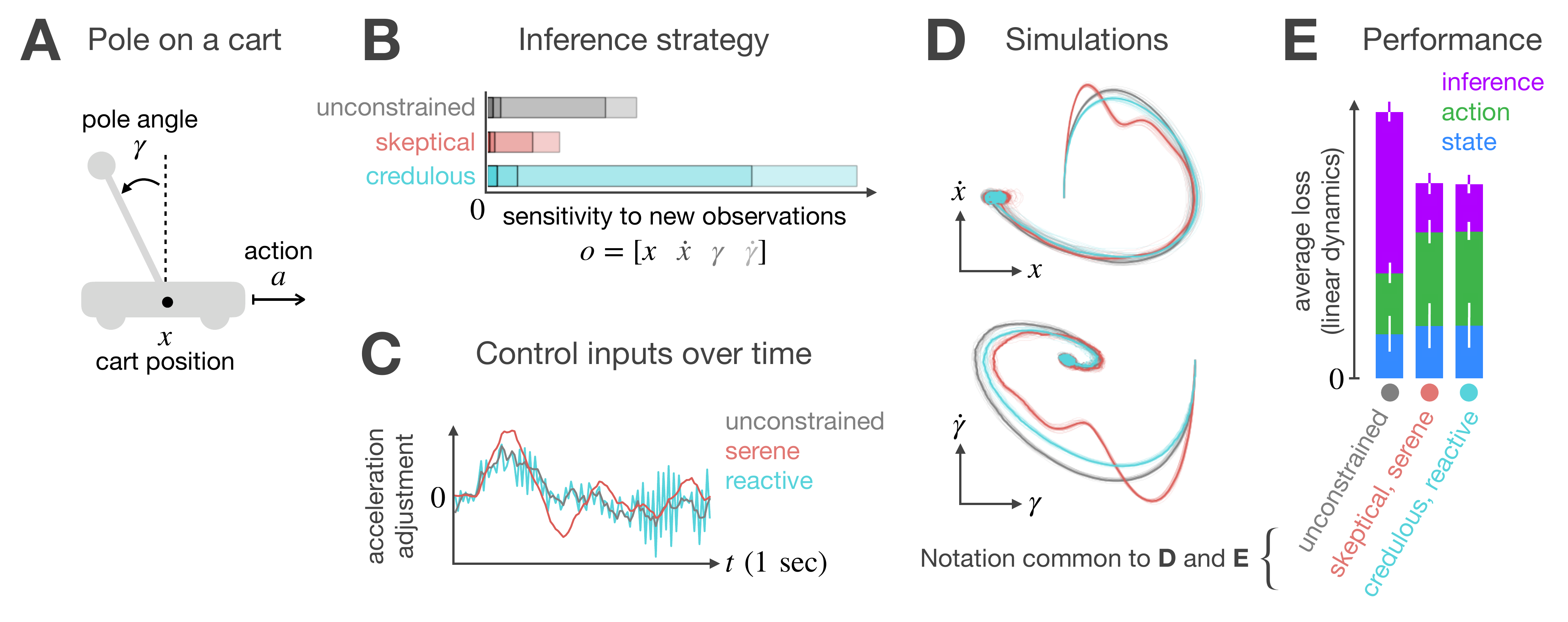}
\caption{Frugal control for balancing a pole. \textbf{A)} Schematic of relevant variables. The controller aims to balance the pole on a moving cart by adjusting the cart's acceleration. \textbf{B)} Inference sensitivity to new observations. Color intensity indicate weighting of individual observation dimensions. The unconstrained agent (gray) weighs observations based solely on statistical reliability. In contrast, frugal agents (non-gray) also take control objectives into account, becoming selectively more skeptical or credulous. This strategic reweighting of observations, when paired with an appropriate control policy, reduces inference costs while preserving eventual goal attainment. \textbf{C)} Control trajectories for different agents. Skeptical inference can be compensated by a serene controller that adjusts the cart's acceleration gradually. However, credulous inference requires a reactive controller that frenetically changes the direction of motion. \textbf{D)} State-space trajectories. Both frugal agents (non-gray trajectories) are able to attain the goal, stabilizing the pole at the upright position. Individual trials are displayed in light colors, with the mean trajectory emphasized in dark.  \textbf{E)} Statistical performance at equilibrium. Both frugal strategies achieve statistically indistinguishable state, action, and inference costs under the linear approximation, demonstrating that they represent qualitatively distinct yet equally effective solutions to the computationally constrained control problem. Error bars indicate one standard deviation}
\label{fig:F5}
\end{figure*}

Our first task is a classic control problem: balancing a pole on a moving cart (Figure \ref{fig:F5}A). To balance the pole, the cart moves forward and backward. The action space is one-dimensional, with actions controlling the cart's acceleration. The hidden state is a four-dimensional vector  $(x, \dot{x}, \gamma, \dot{\gamma})$, representing the cart’s position, velocity, pole angle, and angular velocity. Instead of observing the true state directly, the agent receives a four-dimensional noisy observation vector. Noise is present in all state variables, though it is more pronounced in the velocity and the angular velocity. A real-world application of this problem is the Segway—a self-balancing transportation device that allows tourists to explore cities while avoiding traffic jams. As with many mass-market intelligent devices, the Segway must support high-frequency control under limited on-chip computational resources, making efficiency in sensing and inference a practical concern.

For an agent operating in a one-dimensional action space and prioritizing saving bits in the inference (high $C_b$), the solution to the planning problem is a family of two frugal strategies. We examine the behavior of agents implementing those strategies and compare their performance to that of an unconstrained agent ($C_b =0$). When integrating new evidence (Figure \ref{fig:F5}B), the unconstrained agent adopts a purely statistical approach, weighting each dimension of the observation in proportion to its reliability. In contrast, frugal agents take into account both measurement reliability and control objectives. This allows saving bits in the inference by strategically adjusting observation weights and control gain while minimally compromising eventual goal attainment. Figure \ref{fig:F5}C illustrates the two possible mechanisms that help mitigate the estimation errors arising from frugal inference. Errors that result from attenuating observation weights (skeptical inference) are counteracted by smooth, gradual adjustments to the cart's acceleration (serene control). However, offsetting the errors caused by amplifying observation weights (credulous inference) requires frenetic adjustments to the direction of motion (reactive control). Figure \ref{fig:F5}D illustrates the state-space trajectories that each strategy induces. Although the frugal strategies differ noticeably during the transient, both successfully drive the system to an equilibrium near the target state, keeping the pole upright. Statistical analysis of performance at equilibrium under the approximated linear model (Figure \ref{fig:F5}E) confirms that both frugal strategies solve the computationally constrained control problem equally well, achieving statistically equivalent state, action, and inference costs. Although this equivalence may seem counterintuitive, given that one strategy produces smooth actions and the other exhibits frenetic fluctuations, the action cost remains identical because the loss function penalizes total control magnitude, not temporal variability. Thus, despite differing trajectories, the cumulative action cost is equivalent under our evaluation metric. When evaluated under the original nonlinear dynamics, performance differs slightly due to model mismatch; however, goal attainment remains unaffected, as illustrated in Figure \ref{fig:F5}D.

\subsubsection*{Frugal drone}

\begin{figure*}[t!]
\centering
\includegraphics[scale=0.25]{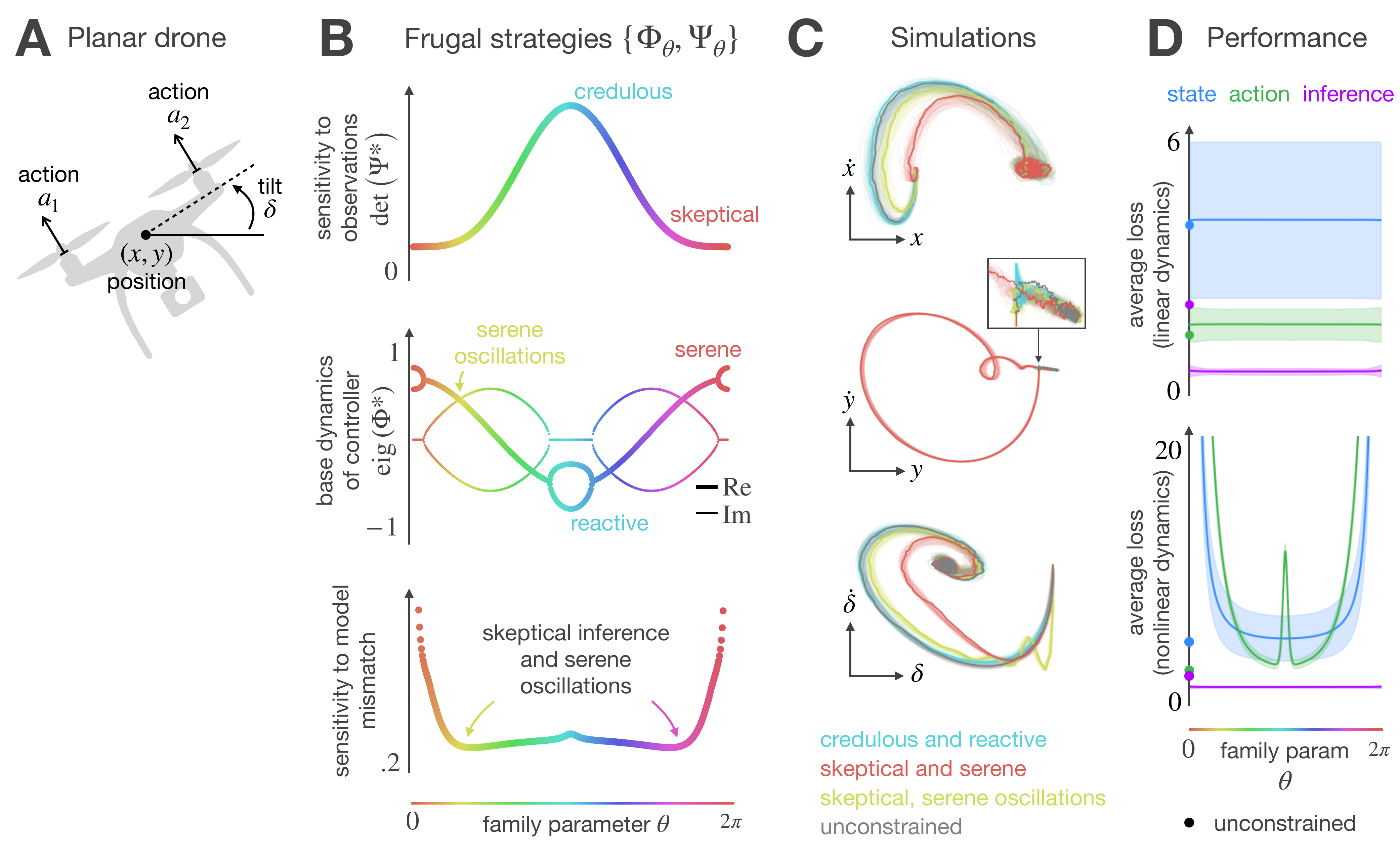}
\caption{Frugal control of a planar drone maintaining a fixed hover position. \textbf{A)} Schematic of relevant variables. \textbf{B)} Family of frugal strategies. For a two-dimensional controller, the solution to the planning problem comprises infinite combinations of lossy inference and error-aware control. These strategies differ in how the agent integrates new evidence (top), offsets estimation errors (middle), and generalizes to novel settings (bottom).  \textbf{C)} State-space trajectories. The frugal strategies successfully drive the system to an equilibrium near the target state. During the transient, combining skeptical inference with serene control yields state-space trajectories that differ substantially from those generated by an unconstrained agent, a credulous and reactive agent, and a serene agent with oscillations. This behavior is in line with our sensitivity analysis (panel B, bottom), which indicates that combining skeptical inference with serene control is highly sensitive to model mismatch. Here, the mismatch arises because the trajectories reflect the true nonlinear dynamics, whereas the strategies were computed using linearized approximations of those dynamics. Individual trials are displayed in light colors, with the mean trajectory emphasized in dark. \textbf{D)} Statistical performance at equilibrium. All family members perform equally well under linear dynamics (top), but respond differently when evaluated using simulations of the nonlinear model (bottom). Mean cost is shown by lines, with shaded regions denoting one standard deviation; unconstrained performance is marked by a dot on the vertical axis.}
\label{fig:F6}
\end{figure*}

Here we assess a drone's ability to maintain a stable hover in the presence of gravity and external disturbances such as wind gusts. The drone is restricted to move within the $xy$-plane and is modeled as a rigid body equipped with two propellers (Figure \ref{fig:F6}A).  The action space is two-dimensional: thrust commands to the propellers that jointly control the drone's altitude and orientation. The hidden state is a six-dimensional vector $(x, y, \delta, \dot{x}, \dot{y}, \dot{\delta})$, representing the drone's horizontal and vertical positions, tilt angle, and their respective velocities. The agent receives a six-dimensional observation vector subject to noise in all state variables, with the greatest impact on velocity components. Autonomous drones operate under tight energy and size-, weight-, and power-constrained onboard computation, leaving limited processing capacity for compute-intensive inference.

For an agent operating in a two-dimensional action space and prioritizing saving bits in the inference (high $C_b$), the solution to the planning problem is an infinite set of frugal strategies (Figure \ref{fig:F6}B). Although these multiple combinations of lossy inference and error-aware control achieve statistically equivalent performance under the linear model for which they were optimized, they respond differently to model perturbations. To examine their sensitivity to model mismatch, we introduce subtle variations in the drone's mass and arm length—parameters that directly influence the motor responsiveness. We quantify robustness by computing the natural gradient of the expected loss with respect to these parameters, capturing worst-case local sensitivity to model mismatch. As shown in Figure \ref{fig:F6}B-bottom, frugal strategies differ markedly in robustness. In particular, the combination of skeptical inference with serene control is highly sensitive to model mismatch, whereas a slight modification that introduces oscillations in the controller’s base dynamics yields the most robust behavior across the family. Importantly, this improvement does not require re-optimizing the original objective, but instead arises from moving within the structured family of equally effective strategies, highlighting the advantage of a free design subspace that can accommodate additional constraints, such as robustness, without compromising optimal performance.

Figure \ref{fig:F6}C compares the state-space trajectories generated by three strategies in the solution family: $\Pi_1$, which combines credulous inference and reactive control, $\Pi_2$, which pairs skeptical inference with serene control, and $\Pi_3$, which combines skeptical inference and smooth oscillating control. As expected, frugal strategies generate noticeably different trajectories during the transient. The strategy $\Pi_1$, which relies on frenetic corrections to estimation errors, induces trajectories that closely resemble those of an unconstrained agent ($C_b = 0$). The resemblance is convenient, but the frenetic corrections may increase actuator wear. The strategy $\Pi_2$, which compensates for estimation errors through smooth and gradual adjustments, is better suited to reduce actuator wear. However, smooth control comes at the expense of larger deviations from unconstrained performance. Interestingly, strategy $\Pi_3$ strikes a balance between the two: it achieves state trajectories that resemble those of an unconstrained agent while relying on smooth, oscillatory actions that may reduce stress on hardware during implementation. This result further illustrates how the structured family of frugal strategies enables systematic adjustments that accommodate objectives not considered during the original optimization, such as accuracy and smoothness, without sacrificing optimality on the original task. Finally, Figure \ref{fig:F6}D presents the performance at equilibrium of the complete solution family. 

\section*{Discussion}\label{secDis}

We introduced a novel variant of the POMDP framework in which the information distilled from previous evidence is treated as a resource that the agent can directly regulate. This adds a new dimension to the classic trade-off between goal achievement and physical effort: the internal cost of resolving state uncertainty. In multivariate settings, this perspective generalizes the principle of minimal intervention—agents act not only on deviations that impair performance, but also selectively regulate which state dimensions must be inferred accurately and which can be safely attenuated. Treating inference as a regulated process rather than a fixed subroutine is a key departure from existing information-constrained decision-making frameworks, in which control does not explicitly modulate the intrinsic complexity of the inference problem. To elucidate the structure of this new regime, we solve a local linear–quadratic–Gaussian (LQG) approximation of the general problem, revealing three principles of computational efficiency. First, frugal agents engage in Bayes-optimal (lossless) inference only when it is necessary or affordable; otherwise, they leave some epistemic uncertainty unresolved and compensate through control. Second, beyond this transition, the solution is a structured family of equally effective strategies that differ in how agents coordinate internal computations with external movement. This multiplicity enables flexibility, as frugal agents can adapt to new objectives or constraints by moving within this solution family without sacrificing performance on the original task. Third, frugal agents systematically trade control effort for reductions in inference cost, increasing movement either to reduce state variability that inference would otherwise need to resolve or to counteract estimation errors arising from imperfect inference. Although derived under a local LQG approximation, these principles persist in nonlinear systems. Empirical results on tasks such as pole balancing and drone stabilization show that frugal strategies remain effective, suggesting that the observed behaviors reflect fundamental consequences of accounting for inference costs rather than artifacts of linearization.

While our results advance the understanding of how the computational burden of inference reshapes rational behavior in complex, partially observed control tasks, we acknowledge several important caveats. We adopt a linear–quadratic-Gaussian approximation to the underlying dynamics, assuming that such models provide a reasonable local representation of the true process. This assumption has repeatedly yielded foundational insights in control and learning \cite{todorov2006linearly, tang2023analysis, hu2023toward, todorov2005stochastic, susemihl2014optimal, boominathan2022phase}, and has enabled solving complex, nonlinear, continuous-state control problems that would otherwise be intractable \cite{levine2014learning, gao2025enabling}. As in this prior work, the approximation enabled a precise characterization of how new control structures emerge when inference itself becomes a regulated, costly mechanism. However, this simplification imposes clear limitations. We assumed jointly Gaussian process and observation noise with covariances independent of controls, world states \cite{todorov2005stochastic}, and inference states, and we did not model internal computational noise \cite{boominathan2022phase}. Moreover, there exist important regimes in which local linear–quadratic-Gaussian approximations break down. Systems with strong nonlinearities and discontinuities—such as contact-rich manipulation and legged locomotion with intermittent ground contact—may undergo rapid transitions across qualitatively distinct behaviors that invalidate a single local approximation. High-dimensional sensory inputs, including raw visual observations, further compound these challenges by introducing nonlinear observation mappings and complex latent structure. Several extensions may help mitigate these limitations. Piecewise-linear or hybrid dynamical models, iterative local re-linearization along trajectories, and learned latent representations provide structured approaches for broadening the applicability of our framework. Relaxing our assumptions will, in general, preclude closed-form solutions. Nevertheless, reinforcement learning and approximate planning methods offer practical alternatives. While the resulting policies may no longer admit analytical characterization, they can be evaluated empirically, thereby enabling systematic assessment of the qualitative predictions developed here.

Our results are particularly relevant as resource-efficient intelligence becomes a central challenge for autonomous systems. Robots such as NASA’s Valkyrie and Boston Dynamics’ Atlas demonstrate that progress toward scalable autonomy is increasingly constrained not by mechanics, but by the computational demands of robust reasoning \cite{wensing2023optimization}. Achieving full autonomy in realistic, untethered settings requires algorithms capable of processing, integrating, and acting on high-dimensional, multi-modal, and noisy sensory data, all while operating under strict limits on onboard computation and energy availability \cite{gu2025humanoid}. As historical gains in computing power slow and sensory complexity continues to rise, it is no longer tenable to assume that inference can be performed optimally—or even near optimally—whenever uncertainty arises. Instead, inference itself must be treated as a regulated and costly component of the decision-making process. This perspective departs from the traditional modular separation of inference and control, but closely mirrors biological intelligence, where perception, inference, and control are tightly coordinated to manage severe resource constraints. Living organisms reliably transform noisy sensory inputs into adaptive behavior across diverse tasks using limited experience, modest processing capacity, and energy budgets orders of magnitude below those of modern machines \cite{padamsey2022neocortex, kashiri2018overview}. This efficiency is supported by neural mechanisms that efficiently encode sensory information \cite{attneave1954some, barlow1961possible, pitkow2012decorrelation, wei2015bayesian, zheng2025unbearable} and flexibly reallocate cognitive resources as task demands change \cite{laughlin1981simple, vul2014one, ho2022people, tavoni2022human, fang2024sequential}. Accordingly, beyond its engineering relevance, our normative framework establishes a principled bridge between formal theories of optimal control and empirical observations in biological systems. By formulating inference costs in information-theoretic—and ultimately neural—terms, our framework suggests a route for linking abstract notions of computational effort to concrete control policies, opening the door to experimentally testable connections between behavior, control structure, and neural activity. One potential signature of frugal inference is the emergence of structured behavioral diversity: agents facing identical task demands may adopt systematically different yet equally effective strategies, reflecting alternative allocations of internal resources rather than noise or suboptimality. Extending our frugal POMDP framework to nonstationary environments, incorporating internal computational noise, state- and action-dependent variability, and relaxing linear-Gaussian assumptions on belief representations will be necessary to capture the richness of biological systems and natural tasks. Nonetheless, the present results already establish a principled account of how inference and control can be jointly optimized under resource constraints in partially observed environments, providing a foundation for a new type of rational computation that supports effective yet resource-efficient control in both biological and artificial systems.

\section*{Methods} \label{sec:secMet}

Here we outline our approach to formulate, solve, and analyze frugal POMDPs. We describe the modeling assumptions, numerical optimization procedures, and analytical tools that reveal the structure of the solution family, enabling a principled study of resource–utility trade-offs when inference and control are jointly optimized.

\subsection*{Problem formulation}

In our frugal variant of the POMDP framework, belief updating is no longer a fixed subroutine, but a cost-sensitive process optimized jointly with control (action selection). In general settings, the parameters defining the solution to this computationally constrained control problem can be complex, such as the weights of a recurrent neural network. However, when the dynamics are locally approximated by a linear-Gaussian model and the reward function by a quadratic form, the problem becomes tractable and the solution interpretable. Under this approximation, the hidden state evolves according to stochastic linear dynamics, $s_t = D s_{t-1} + E a_{t-1} + w_{t-1}$, and observations are linear, noisy versions of the hidden state $o_t = s_t + v_t$. Here, the dynamics matrix $D $ captures how unstable the state is, the input gain matrix $E $ characterizes actuator responsiveness, $a_{t-1} $ is the action taken by the agent, $w_{t-1} $ is additive white Gaussian noise with isotropic covariance $Q $, and $v_t $ is additive white Gaussian noise with isotropic covariance $R $. Thus, finding the frugal strategy $\Pi=\left\{\pi^b, \pi^a\right\}$ that balances goal achievement, motion effort, and the information gained through inference boils down to solving the following optimization problem:
\begin{equation}
\min_{\left\{\pi^b,  \pi^a\right\}} \mathbb{E}_{\tau \sim p(\tau)
} \left[ \sum_t s_t^\top C_s s_t  + a_t^\top C_a a_t  + C_b \ \mathcal{I}\left(s_t; b_t\right) \right] \ \\
\label{eq:opt_prob}
\end{equation}
\noindent subject to:
\begin{align*}
s_t & = Ds_{t-1} + Ea_{t-1} + w_{t-1} \ \ ; \ \ w_{t-1} \sim \mathcal{N}(\mathbf{0}, Q)\\
o_t & = s_{t} + v_{t} \ \ ; \ \ v_{t} \sim \mathcal{N}(\mathbf{0}, R)\\
b_t & = f\left(o_{
\leq t}, a_{<t} ; \pi^b\right) \\
a_t & = g\left(b_{t} ; \pi^a\right)
\end{align*}
\noindent where $f(\cdot)$ defines the inference process that integrates new evidence under parameters $\pi^b$, and $g(\cdot)$ specifies the control policy that maps the resulting beliefs to actions under parameters $\pi^a$. In the cost function, the penalties $\Xi = \{C_s, C_a, C_b\}$ determine the relative importance of the competing objectives: minimizing state deviations, reducing motion effort, and saving bits in the inference. Crucially, the expectation $\mathbb{E}$ is taken with respect to the probability distribution of trajectories $\tau=\{s_{0:T}, o_{0:T}, a_{0:T}, b_{0:T}\}$ that the frugal strategy generates given the dynamics defined by the world properties $\Omega=\left\{D, E, Q, R\right\}$. We assume that both $\Omega$ and $\Xi$ are known and change slowly. This enables the agent to decide how to compress previous evidence while observing the real consequences of its actions, and compensate for estimation errors that result from cheaper inference through additional motion effort.

\subsection*{Parameterizing the solution}

Exact belief updating follows directly from recursive Bayesian inference:
\begin{align*}
b_t &= p(s_t | o_0, \cdots, o_t, a_0, \cdots, a_{t-1}) \\
& \propto p(o_t | s_t) \int p(s_t | s_{t-1}, a_{t-1}) b_{t-1} {\rm d}s_{t-1}
\end{align*}
In this process, the belief is propagated through a transition model $p(s_t | s_{t-1}, a_{t-1})$ and updated using an observation model $p(o_t | s_t)$ via Bayes rule. In POMDPs with linear-Gaussian dynamics, this process is analytically tractable and yields a Gaussian posterior:  $b_t = \mathcal{N}(\hat{s}_t, P_t)$. The closed-form expressions for the posterior mean and covariance are:
\begin{align}
\hat{s}_t
&= \mathcal{D}\hat{s}_{t-1} + \mathcal{E}a_{t-1}
   + \beta\!\left(o_t - \bigl(\mathcal{D}\hat{s}_{t-1} + \mathcal{E}a_{t-1}\bigr)\right)
\label{eq:mean_exp}
\\
P_t
&= (\mathbf{I}-\beta)\bigl(\mathcal{D} P_{t-1}\mathcal{D}^\top + \mathcal{Q}\bigr)
\label{eq:cov_exp}
\end{align}
\noindent with $\beta =
\bigl(\mathcal{D} P_{t-1} \mathcal{D}^\top + \mathcal{Q}\bigr)
\bigl(\mathcal{D} P_{t-1} \mathcal{D}^\top + \mathcal{Q} + \mathcal{R}\bigr)^{-1}$. Here, $\tilde{\Omega}=\{\mathcal{D}, \mathcal{E}, \mathcal{Q}, \mathcal{R}\}$ is the set of parameters describing how the agent assumes the hidden state evolves and generates observations. These closed-form expressions define the celebrated Kalman filter, and yield an exact posterior distribution when $\tilde{\Omega}$ faithfully represents reality. Our agents capitalize on this analytical tractability but can tune the parameters $\tilde{\Omega}$ to modulate inference quality.

Equations \ref{eq:mean_exp} and \ref{eq:cov_exp} admit further simplification. When the parameters $\tilde{\Omega}$ are time-invariant, the posterior covariance converges to a steady-state value $P$. Upon convergence—and assuming actions are linear functions of the state estimate: $a_t = L \hat{s}_t$—Equation \ref{eq:mean_exp} takes the form of an exponential filter:
\begin{align*}
\hat{s}_t &=  \left(\mathcal{D}  + \mathcal{E} L \right) \hat{s}_{t-1} + \beta \left(o_t -\left(\mathcal{D}  + \mathcal{E} L \right) \hat{s}_{t-1}\right) \\
&=  \left[\left(\mathcal{D}  + \mathcal{E} L \right)\left(\mathbf{I} - \beta \right)\right] \hat{s}_{t-1} + \beta o_t  \\
&=  \Gamma \hat{s}_{t-1} + \beta o_t  \\
& = \beta \sum_{i=0}^t \Gamma^i o_{t-i} 
\end{align*}
Therefore, for a frugal POMDP locally approximated by linear-Gaussian dynamics, the parameters $\{\Gamma, \beta\}=\pi^b$ completely define the inference process. Here, $\Gamma$ indicates how much of the past should be remembered, while $\beta$ scales observations to minimize estimation bias. We assume that actions are linear functions of the state estimate: $a_t = L \hat{s}_t$; thus, the control policy is fully parameterized by the control gain $\{L\}=\pi^a$. We refer to the parameters $\{\Gamma, \beta, L\}$ that optimize problem \ref{eq:opt_prob} as the {\em frugal strategy}.

\subsection*{Computing frugal strategies}

While we exploit the linear-Gaussian structure of the problem to compute and interpret frugal strategies, our approach differs from classic LQG control. In our frugal POMDP, the agent pays for every bit of information gained through inference, with a cost rate modulated by the parameter $C_b$ in the loss function. This incentivizes the agent to jointly optimize inference and control, which poses a challenge for the conventional LQG controller. When the belief fails to fully capture the history of past observations and actions, it cannot restore the Markov property that the past and the future are conditionally independent given the present. Since the LQG controller relies on this property to guarantee the optimality of its solutions, modulating inference quality undermines its effectiveness. Finding the strategy that solves a frugal POMDP thus requires reasoning over a joint space of states and actions. To address this challenge, we create an augmented state variable $z_t=[s_t, a_t]^\top$ that describes the joint evolution of states and actions:
\begin{align*}
        z_t &= 
        \begin{bmatrix}
            D & E \\
            D \Psi & \Pi + E \Psi
        \end{bmatrix}
        z_{t-1} + 
        \begin{bmatrix}
            w_{t-1} \\
            \Psi(w_{t-1} + v_t)
        \end{bmatrix} \\
        & = \mathrm{M} z_{t-1} + \eta_{t-1}    \ ; \ \eta_{t-1} \sim \mathcal{N}(\mathbf{0}, \Upsilon)
\end{align*}
Here, $M$ describes the base dynamics of the augmented state $z_t$, $\Upsilon$ characterizes the randomness in the joint space of states and actions, and $\{\Phi=L \Gamma L^{+}, \Psi=L \beta\}$  represent the parameters of the controller's input-output form: $a_t = \Phi a_{t-1} + \Psi o_t$. Here, the controller's base dynamics $\Phi$ defines how the actions evolve without new evidence, the observation sensitivity $\Psi$ determines which dimensions can be attenuated or magnified without compromising eventual goal attainment, and the symbol $^+$ denotes the Moore–Penrose pseudoinverse. If $z_t$ can be stabilized by tuning the parameters  $\{\Phi, \Psi\}$, its probability distribution reaches equilibrium and becomes $p(z_t) =\mathcal{N}(\mathbf{0}, \Sigma)$ for all $t$. Therefore, at equilibrium, the entries of the steady-state covariance matrix $\Sigma = \left[\begin{smallmatrix} \Sigma_s & \Sigma_{as} \\ \Sigma_{sa} & \Sigma_a \end{smallmatrix}\right]$ fully define the components of the loss function. As a result, at equilibrium, our computationally constrained control problem becomes:
\begin{equation}
\min_{\{\Phi, \Psi\}} \left\{  \mathrm{Tr} \left( C_s \Sigma_s + C_a \Sigma_a \right) + C_b \frac{1}{2} \log_2 \frac{\det\left(\Sigma_s\right) \det\left(\Sigma_a\right)}{\det\left(\Sigma\right)} \right\}
\label{eq:opt_prob_eq}
\end{equation}
We solve problem \ref{eq:opt_prob_eq} using stochastic gradient descent, which iteratively adjusts the parameters $\{\Phi, \Psi\}$ to minimize the loss. We verify that the candidate solutions produce a positive definite covariance matrix  $\Sigma$ and a transition matrix $\mathrm{M}$ with stable eigenvalues throughout the optimization process. These conditions guarantee that the trajectory distribution of states and actions is well-defined and that the system can reach the target state. Additionally, we monitor the Hessian of the objective function to ensure that the solutions are locally optimal, stable, and meaningful. The numerical optimization method described here produces the landscapes shown in Figures \ref{fig:F2}B and \ref{fig:F2}C. While those results correspond to a scalar task, the approach generalizes to multivariate problems, as demonstrated by the illustrative tasks in Subsection \textit{Empirical validation in nonlinear environments}.

\subsection*{Recovering the complete solution family}
\label{subsecFamily}

The frugal strategy $\{\Phi^*, \Psi^*\}$ that solves problem \ref{eq:opt_prob_eq} induces a unique covariance matrix $\Sigma^*$. However, the reverse is not true: a given matrix $\Sigma^*$ may correspond to multiple combinations of $\Phi^*$ and $\Psi^*$. To understand the conditions under which this occurs, we analyze the structure of the discrete-time Lyapunov equation that $\Sigma^*$ satisfies:
\begin{align}
\left[\begin{smallmatrix}
\Sigma_s^* & \Sigma_{as}^* \\
\Sigma_{sa}^* & \Sigma_a^*
\end{smallmatrix}\right]
&=
\left[\begin{smallmatrix}
D & E \\
D\Psi^* & \Phi^* + E\Psi^*
\end{smallmatrix}\right]
\left[\begin{smallmatrix}
\Sigma_s^* & \Sigma_{as}^* \\
\Sigma_{sa}^* & \Sigma_a^*
\end{smallmatrix}\right]
\left[\begin{smallmatrix}
D & E \\
D\Psi^* & \Phi^* + E\Psi^*
\end{smallmatrix}\right] ^{\!\top}
\notag\\
&\quad+
\left[\begin{smallmatrix}
Q & Q\Psi^{*\top} \\
\Psi^* Q & \Psi^*(Q+R)\Psi^{*\top}
\end{smallmatrix}\right]
\label{eq:aug_fam}
\end{align}
Solving equation \ref{eq:aug_fam} element-wise shows that the frugal strategy $\{\Phi^*, \Psi^*\}$  satisfies a generalized ellipsoidal constraint:
\begin{equation}
    \Sigma_a^* = \Phi^* \Sigma_a^* \Phi^{*\top} + \Psi^* \Sigma_{sa}^{*\top} +\Sigma_{sa}^* \Psi^{*\top} +\Psi^* (R - \Sigma_s^*) \Psi^{*\top}
    \label{eq:ellipse}
\end{equation}
with $\Psi^* = \left(\Sigma_{sa}^* - \Phi^*\left(\Sigma_{sa}^* D^\top + \Sigma_a^* E^\top\right)\right)\Sigma_s^{^*-1}$  and $R \geq \Sigma_s^*$ . Consequently, finding $\Phi^*$ as a function of $\Sigma^*$ and $\Psi^*$ requires solving a quadratic form:
\begin{equation*}
    \Phi^* F_2 \Phi^{*\top} + \Phi F_1 + F_1^\top \Phi^{*\top}=F_0
\end{equation*}
here, $F_0$, $F_1$, and $F_2$ are functions of $\Sigma^*$ and the world properties $\{D, E, Q, R\}$. This quadratic form can be rearranged as:
\begin{equation}
    \left(\Phi^* + F_1^\top F_2 ^{-1}\right)F_2 \left(\Phi^* + F_1^\top F_2 ^{-1}\right)^\top=F_0 + F_1^\top F_2 ^{-\top}F_1
    \label{eq:quadpi}
\end{equation}
The right side of equation \ref{eq:quadpi} quantifies how much epistemic uncertainty remains after updating the belief $b_t=\mathcal{N}(\hat{s}_t, P)$ with the most recent observation $o_t$. That is:
\begin{align}
F_0 + & F_1^\top F_2 ^{-\top}F_1 \propto \ 
{\rm Cov}(s_t - \langle s_t|\hat{s}_t\rangle)\, {\rm Cov}(\langle s_t|\hat{s}_t\rangle)^{-1} \nonumber \\
&- {\rm Cov}(s_{t+1} - \langle s_{t+1}|\hat{s}_t, o_{t+1}\rangle)\,{\rm Cov}(\langle s_{t+1}|\hat{s}_t, o_{t+1}\rangle)^{-1}
\label{eq:lossless_update}
\end{align}
\noindent with $\langle \cdot\rangle$ denoting the expectation taken with respect to the probability distribution of trajectories that the frugal strategy generates given the dynamics defined by the true world properties $\Omega=\left\{D, E, Q, R\right\}$. Equation \ref{eq:lossless_update} quantifies the difference between the uncertainty that remains unexplained about $s_t$ given its estimate $\hat{s}_t =\int s_t\, b(s_t)  {\rm d} s_t$ and the uncertainty that remains unexplained about the next state given the current estimate and the next observation. When this difference is zero, equation \ref{eq:quadpi} takes a linear form and the solution is unique: $\Phi^*=- F_1^\top F_2 ^{-1}$. However, when the agent pays for every bit of information gained through inference, the cost of mitigating epistemic uncertainty may outweigh its benefits in solving the task. When this happens, the unresolved uncertainty yields a slack in the ellipsoidal constraint (Equation \ref{eq:ellipse}) that a frugal strategy must satisfy to balance state, action, and inference costs. This additional freedom gives rise to multiple combinations of imperfect inference and error-aware control that can solve the planning problem equally well.

To recover the complete solution family that emerges when epistemic uncertainty remains unresolved, we calculate the eigenvalue decomposition of $F_2$ on the left side of equation  \ref{eq:quadpi}  and the eigenvalue decomposition of $\xi=F_0 + F_1^\top F_2 ^{-\top}F_1$ on the right side:
\begin{align*}
    \left(\Phi^* + F_1^\top F_2 ^{-1}\right)U_{F_2} \Lambda_{F_2}^\frac{1}{2} \Lambda_{F_2}^\frac{1}{2} U_{F_2}^\top \left(\Phi^* + F_1^\top F_2 ^{-1}\right)^\top  = \\ U_\xi \Lambda_\xi^\frac{1}{2} \Lambda_\xi^\frac{1}{2} U_\xi^\top
\end{align*}
This decomposition reveals a free orthogonal transformation $\Theta$ in the joint space of states and actions:
\begin{align*}
    \left(\Phi^* + F_1^\top F_2 ^{-1}\right)U_{F_2} \Lambda_{F_2}^\frac{1}{2} \Lambda_{F_2}^\frac{1}{2} U_{F_2}^\top \left(\Phi^* + F_1^\top F_2 ^{-1}\right)^\top  = \\ U_\xi \Lambda_\xi^\frac{1}{2} \Theta \Theta^\top\Lambda_\xi^\frac{1}{2} U_\xi^\top
\end{align*}
We can use this transformation to parameterize the frugal strategies in the solution family. Since $\left(\Phi^* + F_1^\top F_2 ^{-1}\right)U_{F_2} \Lambda_{F_2}^\frac{1}{2} =U_\xi \Lambda_\xi^\frac{1}{2} \Theta$, each member of the solution family is specified by:
\begin{align*}
    \Phi_\Theta^*   &=U_\xi \Lambda_\xi^{\frac{1}{2}} \Theta  \Lambda_{F_2}^{-\frac{1}{2}} U_{F_2}^{-1} - F_1^\top F_2 ^{-1} \\
    \Psi_\Theta^* &= \left(\Sigma_{sa} - \Phi_\Theta^*\left(\Sigma_{sa} D^\top + \Sigma_a E^\top\right)\right)\Sigma_s^{-1}
\end{align*}
By construction, each family member $\{\Phi_\Theta^*, \Psi_\Theta^*\}$ satisfies the ellipsoidal constraint (Equation \ref{eq:ellipse}) and, thus, results in the same bounded optimal covariance matrix $\Sigma^*$ whose entries fully define the loss function in problem  \ref{eq:opt_prob_eq}. This explains why all family members perform equally well given the properties $\{D, E, Q, R, C_s, C_a, C_b\}$ for which they are optimized. However, due to differences in their temporal structure, the solutions vary in how the agent models the world to integrate new evidence, offsets estimation errors, and generalizes to novel settings. For high-level implications, see Subsection \textit{Adaptability begins when perfection ends}, where we outline intuition and applications of this mathematical analysis.

\subsection*{Interpreting frugal strategies}\label{subsecInterpreting}

The first step to interpret a frugal strategy  $\{\Phi=L \Gamma L^{+}, \Psi=L \beta\}$ is to recover the parameters of the inference. We recover the filter's memory factor using $\Gamma=(\beta \Psi^+)\Phi (\Psi \beta^{+})$. The observation scaling factor $\beta$ is a degenerate parameter because its effects can be neutralized by the control gain $L$. To address this degeneracy, we let $\beta$ be the value that minimizes estimation bias. The next step is to identify the generative model that shapes the belief $b_t = \mathcal{N}\left(\hat{s}_t, P ; \Gamma, \beta\right)$. Our agents engage in recursive Bayesian inference but distort the generative model to modulate inference quality; thus, their subjective posterior covariance has the following closed-form expression:
\begin{equation}
        P  = \left(\mathbf{I} - \beta\right)\left (\mathcal{D} P  \mathcal{D}^\top + \mathcal{Q}\right)   \label{eq:model1}
\end{equation}
If the agent's assumptions $\tilde{\Omega} = \{\mathcal{D}, \mathcal{E}, \mathcal{Q}, \mathcal{R}\}$ truly reflected reality, the posterior covariance $P$ would equal the mean squared estimation error: 
\begin{align}
P & = \Sigma_e \nonumber \\
&= \mathbb{E}_{\tau \sim p(\tau)} \left[(s_t - \hat{s}_t)^2 | o_{\leq t}, a_{<t}\right] \nonumber \\
&= \Sigma_{s} - \Sigma_{s\hat{s}} - \Sigma_{s\hat{s}}^\top + \Sigma_{\hat{s}} \nonumber \\
&= \Sigma_{s} - (\beta \Psi^{+})\Sigma_{sa} - \Sigma_{sa}^\top (\beta \Psi^{+})^\top+ (\beta \Psi^{+})\Sigma_{a} (\beta \Psi^{+})^\top \label{eq:model2}
\end{align}
\noindent with the expectation $\mathbb{E}$ taken with respect to the probability distribution of trajectories $\tau=\{s_{0:T}, o_{0:T}, a_{0:T}, b_{0:T}\}$ that obey the dynamics defined by the true world properties $\Omega=\left\{D, E, Q, R\right\}$. 

We use equations \ref{eq:model1} and \ref{eq:model2} to recover $\tilde{\Omega}$. To this end, we treat actuator responsiveness and sensor noise covariance as intrinsic properties of the agent; accordingly, in the assumed world model $\tilde{\Omega}$ they are set equal to their true values, $\mathcal{E} = E$ and $\mathcal{R} = R$. We then derive $\mathcal{D}$ and $\mathcal{Q}$ as follows:
\begin{align*}
    \mathcal{D} &= (\mathbf{I}-\beta)^{-1}\  \Gamma - \mathcal{E}\ L \\
    \mathcal{Q} &= \left(\mathcal{R}^{-1}(\mathcal{R}^{-1} - \Sigma_e)\mathcal{R}^{-1}\right)^{-1}-\mathcal{D}\ \Sigma_e \mathcal{D}^\top - \mathcal{R}
\end{align*}
This derivation allowed us to thoroughly characterize the solutions to problem \ref{eq:opt_prob_eq}, leading to the results presented in Figure \ref{fig:F3}C.

\section*{Acknowledgments}
This work was supported by AFOSR grant FA9550-21-1-0422 to XP in the Cognitive and Computational Neuroscience program, by the National Science Foundation and by DoD OUSD (R \& E) under Cooperative Agreement PHY-2229929 (The NSF AI Institute for Artificial and Natural Intelligence, ARNI) to XP, and by a Fulbright–García Robles scholarship to IOC.

\section*{References}


\bibliography{pnas-sample}

\end{document}